\documentclass[11pt,a4paper]{article}
\usepackage[hyperref]{acl2019}
\usepackage{times}
\usepackage{latexsym}
\usepackage{pythontex} 
\usepackage{graphicx} 
\usepackage{url}
\usepackage{array}
\usepackage{multirow}
\usepackage{booktabs}

\usepackage{amsmath}
\usepackage{amssymb}

\aclfinalcopy 

\title{Unpacking the Resilience of \\ SNLI Contradiction Examples to Attacks}

\author{Chetan Verma
    \texttt{chetan.kumar.verma@utexas.edu}
,   Archit Agarwal
    \texttt{aa2023mscs@utexas.edu}
}

\author{Chetan Verma \\
Department of Computer Science \\
University of Texas at Austin \\
\texttt{chetan.kumar.verma@utexas.edu} \\\And
Archit Agarwal \\
Department of Computer Science \\
University of Texas at Austin \\
\texttt{aa2023mscs@utexas.edu} \\}

\date{}

\begin{document}
\maketitle

\begin{abstract}
  Pre-trained models excel on NLI benchmarks like SNLI and MultiNLI, but their true language understanding remains uncertain. Models trained only on hypotheses and labels achieve high accuracy, indicating reliance on dataset biases and spurious correlations. To explore this issue, we applied the Universal Adversarial Attack to examine the model's vulnerabilities. Our analysis revealed substantial drops in accuracy for the entailment and neutral classes, whereas the contradiction class exhibited a smaller decline. Fine-tuning the model on an augmented dataset with adversarial examples restored its performance to near-baseline levels for both the standard and challenge sets. Our findings highlight the value of adversarial triggers in identifying spurious correlations and improving robustness while providing insights into the resilience of the contradiction class to adversarial attacks. 
\end{abstract}

\section{Introduction}
Natural Language Inference (NLI) is a foundational task in Natural Language Processing (NLP) that evaluate's a model’s natural language understanding (NLU). It involves determining whether a hypothesis is true (Entailment), false (Contradiction) or cannot be determined (Neutral) given a premise \cite{10.1007/11736790_9, 50052253bbcc42a28a6525945815f4b4}. This reasoning ability is critical for mimicking human understanding and supports a wide range of applications. Consequently, when models achieve high accuracy on this task, it is often claimed that they have strong NLU capabilities. However, recent research \cite{DBLP:journals/corr/abs-1805-01042, DBLP:journals/corr/abs-1803-02324} shows that models achieve high accuracy even when trained on hypothesis-only datasets. This suggests that models exploit spurious correlations and superficial patterns known as dataset artifacts to predict the correct label rather than genuinely understanding the language.

To investigate this phenomenon and explore ways to improve the model, we focused on the Stanford NLI (SNLI) dataset \cite{DBLP:journals/corr/BowmanAPM15}, one of the most widely used benchmarks for NLI tasks. For our study, we selected Efficiently Learning an Encoder that Classifies Token Replacements Accurately (ELECTRA-small\footnote{We will use ELECTRA as an alias to refer to ELECTRA-small throughout this paper}) \cite{DBLP:journals/corr/abs-2003-10555}. This model retains the same architecture as Bidirectional Encoder Representations from Transformers (BERT) but incorporates an improved training method. Moreover, ELECTRA is computationally efficient, making it a practical alternative to larger, more resource-intensive models. 

Adversarial datasets were chosen as a key tool for assessing robustness due to their flexibility and cross-dataset applicability. In particular, we explored a method used in a previous study to generate Universal Adversarial Triggers \cite{DBLP:journals/corr/abs-1908-07125} to create a testing dataset and evaluated the robustness of ELECTRA. These triggers are transferable across models for all datasets, and unlike other adversarial attacks, they are context-independent and hence provide new insights into the general input-output patterns learned by the model. 

To address the identified gaps and enhance the model's robustness, we fine-tuned it using a small, augmented training dataset \cite{DBLP:journals/corr/abs-1904-02668}. A detailed discussion of our methodology and analysis is provided later in the paper.

\section{Background and Related Work}
The evaluation and enhancement of machine learning models' robustness has emerged as a critical focus in recent research. A variety of techniques have been developed to create challenge sets that expose model vulnerabilities, with notable approaches including: (1) Contrast Sets: These are manually crafted modifications to test data that introduce small, label-changing alterations while maintaining lexical and syntactic integrity \cite{gardner-etal-2020-evaluating}; (2) Checklist Sets: This approach involves a systematic, task-agnostic framework for testing NLP models. Drawing inspiration from behavioral testing principles in software engineering, it employs diverse test types to evaluate model performance \cite{ribeiro-etal-2020-beyond}; (3) Adversarial Challenge Sets: These datasets are deliberately modified to provoke problematic outputs, revealing critical weaknesses in models \cite{jia-liang-2017-adversarial, DBLP:journals/corr/abs-1908-07125}.

To address the vulnerabilities identified through such challenge sets, researchers have proposed and employed several mitigation techniques, including: (1) Adversarial Data Training: This approach incorporates challenge sets directly into the training process or employs adversarial data augmentation to strengthen model performance against adversarial inputs \cite{DBLP:journals/corr/abs-1904-02668}; (2) Ensemble-based debiasing: Involves learning a biased model that captures dataset-specific clues, and then training a debiased model on the residual errors \cite{he-etal-2019-unlearn}

These techniques collectively represent significant strides in fortifying machine learning models against potential vulnerabilities, ensuring more reliable and robust performance in diverse applications.

\begin{table*}[h!]
\centering
\begin{tabular}{|l|p{10cm}|p{3cm}|}
\hline
\textbf{Gold Label} & \textbf{Input (Premise and Hypothesis)} & \textbf{Predicted Label} \\
\hline

\textbf{Entailment} (0) & 
\textbf{Premise:} A woman in a purple shirt buying food from a street vendor. \newline
\textbf{Hypothesis:} \textcolor{red}{nobody} a woman makes a purchase from a vendor outside. & 
0 $\rightarrow$ 2 \\
\hline

\textbf{Neutral} (1) & 
\textbf{Premise:} A youth is kicking a soccer ball in an empty brick area. \newline
\textbf{Hypothesis:} \textcolor{red}{cats} a funny human kicking. & 
1 $\rightarrow$ 2 \\
\hline

\textbf{Contradiction} (2) & 
\textbf{Premise:} A person dressed in red and black outside a cracked wall. \newline
\textbf{Hypothesis:} \textcolor{red}{anxiously} a person in red and black falling through the ionosphere. & 
2 $\rightarrow$ 1 \\
\hline

\end{tabular}
\caption{Effect of universal triggers (highlighted in \textcolor{red}{red}) on predicted labels across three classes}
\label{tab:universal_trigger_effect}
\end{table*}

\section{Methodology}

\begin{table*}[t!]
\centering
\begin{tabular}{llccccc}
\toprule
 & \multicolumn{3}{c}{\textbf{Universal}} & \multicolumn{3}{c}{\textbf{Random}} \\
\cmidrule(lr){2-4} \cmidrule(lr){5-7}
\textbf{Targeted Class} & \textbf{Trigger} & \textbf{Majority Class} & \textbf{Score} & \textbf{Trigger} & \textbf{Majority Class} & \textbf{Score} \\
\midrule
\textbf{Entailment} (0) & nobody & 2 & 0.96 & diners & 1 & 0.47 \\
 & no & 2 & 0.83 & hands & 2 & 0.38 \\
\textbf{Neutral} (1) & cats & 2 & 0.96 & road & 1 & 0.40 \\
 & cat & 2 & 0.85 & mass & 0 & 0.38 \\
\textbf{Contradiction} (2) & joyously & 1 & 1.00 & remain & 1 & 0.62 \\
 & celebrating & 1 & 0.79 & rose & 1 & 0.59 \\
\bottomrule
\end{tabular}
\caption{Comparison of Universal and Random Triggers}
\label{tab:score_table}
\end{table*}

\subsection{Universal Adversarial Attack}
\label{ssec:Universal}
Universal adversarial triggers are generated tokens designed to manipulate a model's predictions. When appended to the beginning or end of an input, these triggers can compel the model to produce a prediction that deviates from the gold label. 

\textbf{Why Universal Triggers ?} (1) Black-Box Capability: These triggers can be generated without any access to the target model, making them effective even in black-box scenarios; (2) Model Transferability: These triggers are universal, that is, they are capable of attacking and transferring across different models; and (3) Context Independence: Their independence from context provides valuable insights into the general input-output patterns of the model. Table-\ref{tab:universal_trigger_effect} illustrates the impact of universal adversarial triggers in altering predicted labels, highlighting their effectiveness in manipulating model outputs. 

\subsubsection{Attack Equation}
Given a model \( f \) (with white-box access), a text input composed of tokens \( t \) (which could represent words, sub-words, or characters), and a target label \( \tilde{y} \), the goal is to generate triggers \( t_{\text{adv}} \) to append to the front or back of the input token \( t \). 
In a non-universal adversarial setting, this can be mathematically expressed as:
\begin{equation}
    f(t_{\text{adv}}; t) = \tilde{y}
\end{equation}
\par For a universal adversarial setting, the goal is to optimize the universal trigger such that the loss function for the target class \( \tilde{y} \) is minimized across all inputs from a dataset. Mathematically, this can be represented as:
\begin{equation}
    \underset{t_{\text{adv}}}{\arg\min} \mathbb{E}_{t \sim T} \big[ \mathcal{L}(\tilde{y}, f(t_{\text{adv}}; t)) \big]
\end{equation}
where \( T \) represents all input instances from the dataset, and \( \mathcal{L} \) represents the loss function.
 
\subsubsection{Trigger Search Algorithm}
\label{ssec:Algorithm}
Next, we start by selecting a trigger length: longer triggers tend to be effective, whereas shorter triggers are less noticeable. To initialize trigger creation, we prepend a simple token such as the character \textit{a} or word \textit{the} to the beginning of all inputs.

We incrementally refine the tokens in the trigger to optimize the loss function associated with target prediction, leveraging a technique inspired by HotFlip \cite{ebrahimi-etal-2018-hotflip}, that uses the token's gradient to get the token replacement. To use this technique, the trigger token \( t_{\text{adv}} \) is represented as one-hot vectors and embedded to form \( e_{\text{adv}} \).

The HotFlip-inspired technique uses a linear approximation of the task loss. Specifically we update the embeddings for each trigger token \( e_{\text{adv}_i} \) to minimize the loss by applying a first-order Taylor approximation around the current token embedding:
\begin{equation}
\arg\min_{e'_i \in \mathcal{V}} \left[e'_i - e_{\text{adv}_i}\right]^T \nabla e_{\text{adv}_i} \mathcal{L} 
\end{equation}
where \(\mathcal{V}\) is the set of all token embeddings in the model’s vocabulary, and \(\nabla e_{\text{adv}_i} \mathcal{L} \) is the average of the task loss over a batch. For our use-case, NLI (a classification task), we use the cross-entropy loss to optimize the attack.

\subsection{Inoculation by Fine-Tuning}
\label{ssec:Inoculation}
To address the identified vulnerabilities, we employed the \textit{Inoculation by Fine-Tuning} technique \cite{DBLP:journals/corr/abs-1904-02668}. This method involves fine-tuning a pre-trained model on a small, carefully designed training dataset. Following fine-tuning, the model typically exhibits one of three behaviors: 
\begin{enumerate}
    \item \textbf{Reduced Performance Gap}: The performance disparity between the original test set and the challenge set decreases, with the model maintaining strong performance across both datasets. This outcome suggests that the observed gap originates from the dataset itself rather than inherent limitations of the model.
    \item \textbf{Unchanged Performance}: The model's performance remains static, indicating an inability to adapt to the challenge set. This points to potential limitations within the model's architecture or design as the root cause.
    \item \textbf{Decreased Performance}: The model's performance on the original dataset declines, even if improvements are observed on the challenge set. This behavior indicates potential overfitting to the adversarial examples introduced during fine-tuning, rather than addressing the underlying issue.
\end{enumerate}
By applying this technique, we effectively diagnosed and mitigated the identified issues, strengthening the system's resilience and addressing key vulnerabilities.

\section{Experiments}
This section describes the process of generating triggers (a word in this case), creating attacks using these triggers, and subsequently training and evaluating the ELECTRA model to assess its ability to learn and perform the underlying NLI task effectively.

\subsection{Generation of Triggers}

\subsubsection{Universal Triggers}
\label{sssec:Universal_Triggers}
Universal triggers are created using the Universal Adversarial Attack (Section-\ref{ssec:Universal}). Initially, the token representing the word \textit{the} is prepended to the targeted examples and then trigger search algorithm (Section-\ref{ssec:Algorithm}) is applied to generate the universal triggers. These triggers are derived using the Enhanced Sequential Inference Model (ESIM) \cite{Chen_2017} with GloVe embeddings \cite{pennington-etal-2014-glove}. To simulate a realistic scenario, the generated triggers are tested using the ELECTRA model, assuming black-box access. This setup mimics real-world conditions, where white-box access to deployed models is typically unavailable, but testing their robustness is still necessary.

\subsubsection{Random Triggers}
\label{sssec:Random_Triggers}
As a baseline for the Universal Adversarial Attack, triggers are generated using a Random Attack approach. In this method, words are randomly selected from the SNLI dataset's vocabulary, ensuring a uniform distribution across all three classes. These selected words are then prepended to the hypotheses of SNLI examples to create the random attack.

\subsubsection{Examples and Correlation Score}
Table-\ref{tab:score_table} presents examples of both universal and random triggers, along with their corresponding majority class (the class in which the trigger appears most frequently) and correlation score. The correlation score is defined as the conditional probability of a label \( l \) given a word \( w \), and it is mathematically expressed as: 
\begin{equation}
    p(l|w) = \frac{count(w, l)}{count(w)}
\end{equation}

\subsection{Challenge Sets and Trigger-Augmented Dataset}
\label{ssec: Challenge_sets_trigger}

To systematically evaluate model performance and mitigate reliance on spurious correlations, we developed two challenge sets and a Trigger-Augmented dataset. The challenge sets assess the model's robustness under adversarial conditions, while the Trigger-Augmented dataset aims to enhance generalization by addressing dataset-specific biases. The construction and purpose of these datasets are described below:

\begin{enumerate}
\item \textbf{Challenge Set I} (Validation split with universal triggers\footnote{Our database for universal triggers can be found at \url{https://huggingface.co/datasets/ckverma/snli_universal}}): This set evaluates the model's robustness and understanding of the core NLI task. It consists of 1,000 examples, randomly sampled from each label class in the validation split of the SNLI dataset. Universal triggers (detailed in Section-\ref{sssec:Universal_Triggers}) are prepended to the hypothesis in these examples. This setup enables a focused evaluation of the model's performance across all label classes in the presence of adversarial triggers.

\item \textbf{Challenge Set II} (Validation split with random triggers\footnote{Our database for random triggers can be found at \url{https://huggingface.co/datasets/ckverma/snli_random}}): Designed as a baseline for comparison with universal triggers, this set follows the same construction process as Challenge Set I but replaces universal triggers with random triggers (detailed in Section-\ref{sssec:Random_Triggers}). This set provides a point of reference for measuring the impact of universal triggers on model performance by isolating the effect of non-specific, randomly chosen triggers.
	
\item \textbf{Trigger-Augmented Dataset} (Train split with universal triggers): To reduce the model's reliance on spurious correlations present in the original SNLI dataset, a fine-tuning dataset was created. This dataset contains 6,000 training examples, with 3,000 left unmodified and the remaining 3,000 modified by prepending universal triggers to their hypothesis. This augmentation encourages the model to prioritize semantically meaningful features over spurious patterns during training (refer to Section-\ref{ssec:Inoculation}).

\end{enumerate}

By utilizing these datasets, we aim to systematically evaluate and enhance the robustness of the ELECTRA model under both adversarial and standard conditions.

\subsection{Training and Evaluation process}

The ELECTRA model\footnote{Our code for training the model can be found at \url{https://github.com/ckvermaAI/SNLI-Attack-Analysis.git} } was trained and fine-tuned on a single machine equipped with an NVIDIA T4 GPU. The training process utilized the HuggingFace Trainer framework, configured with a maximum sequence length of 128 to ensure that over 96\% of examples from the SNLI dataset were fully captured without truncation. A batch size of 256 was selected, while all other parameters were left at their default settings.  

Training was conducted in two stages. First, the model was trained on the original SNLI dataset for three epochs to establish a strong foundational understanding of natural language inference. This was followed by a fine-tuning phase, where the model was adapted using the Trigger-Augmented dataset to enhance robustness and mitigate reliance on spurious correlations. This two-stage process allowed the model to effectively balance learning from the original data and adapting to the additional challenge sets.

\begin{table*}[h!]
\centering
\resizebox{\textwidth}{!}{%
\begin{tabular}{|l|l|c|c|c|}
\hline
\textbf{Dataset}       & \textbf{Triggers (Model)}                    & \textbf{Entailment (\%)} & \textbf{Neutral (\%)} & \textbf{Contradiction (\%)} \\ \hline
Validation Subset      & Original (Pre-Finetune)                          & 90.23                   & 86.70                 & 91.06                      \\
Challenge Set I        & Universal (Pre-Finetune)       & 25.78                   & 25.76                 & 83.63                      \\ 
Challenge Set II       & Random (Pre-Finetune)       & 71.20                   & 83.98                 & 91.57                      \\ 
Validation Subset      & Original (Post-Finetune)           & 88.92                   & 84.81                 & 91.16                      \\ 
Challenge Set I        & Universal (Post-Finetune)      & 90.13                   & 87.53                 & 91.96                      \\ \hline
\end{tabular}%
}
\caption{Performance Summary of ELECTRA on Validation subset of SNLI dataset and Challenge Sets (I and II)}
\label{tab:summary}
\end{table*}

\section{Results and Analysis}

\begin{table*}[h!]
\centering
\begin{tabular}{l l c c c}
\hline
\textbf{Ground Truth} & \textbf{Data} & \textbf{E\%} & \textbf{N\%} & \textbf{C\%} \\
\hline
\multirow{3}{*}{Entailment} & Validation subset & \textbf{90.23} & 7.65 & 2.11 \\
& Challenge Set I & 25.78 & 32.93 & \textbf{41.29} \\
& Challenge Set II & \textbf{71.20} & 21.85 & 6.95 \\
\hline
\multirow{3}{*}{Neutral} & Validation subset & 7.33 & \textbf{86.70} & 5.97 \\
& Challenge Set I & 0.63 & 25.76 & \textbf{73.61} \\
& Challenge Set II & 5.97 & \textbf{83.98} & 10.05 \\
\hline
\multirow{3}{*}{Contradiction} & Validation subset & 1.31 & 7.63 & \textbf{91.06} \\
& Challenge Set I & 0.60 & 15.76 & \textbf{83.63} \\
& Challenge Set II & 1.41 & 7.03 & \textbf{91.57} \\
\hline
\end{tabular}
\caption{ELECTRA model's prediction distribution for different datasets. Each row shows a particular dataset and each column shows how often model predicts a particular class. For example, on the challenge set I, neutral examples are classified as contradiction examples 73.61\% times.}
\label{tab:confusion_matrix}
\end{table*}

\subsection{Training ELECTRA on the SNLI Dataset}

The Electra model was trained on the SNLI dataset for three epochs, achieving a validation accuracy of 88.98\%. During this initial phase, evaluations were performed on three datasets: the validation subset (comprising 1,000 randomly sampled examples from the SNLI validation split), Challenge Set I, and Challenge Set II. The results from this stage are summarized in the first three rows of Table-\ref{tab:summary}.

\subsection{Fine-Tuning ELECTRA model}

In the second phase, the model underwent fine-tuning on the Trigger-Augmented dataset for one epoch. This step was designed to reduce the model's dependence on spurious correlations present in the original SNLI dataset, thereby enhancing its robustness. Post-fine-tuning, the model was re-evaluated on the validation subset and Challenge Set I, with the results documented in the last two rows of Table-\ref{tab:summary}.

\subsection{Analyzing the Results}

\subsubsection{Effectiveness of Triggers}
Table-\ref{tab:confusion_matrix} highlights the impact of universal and random triggers. Universal triggers effectively alter the model’s predictions for entailment and neutral examples, often shifting them to other classes. In contrast, random triggers have minimal influence, affecting approximately 20\% of the entailment examples. These findings demonstrate the superior efficacy of universal triggers in manipulating model predictions (compared to random triggers).

\subsubsection{Success of Universal Triggers}
The Universal Adversarial Attack generates triggers that are strongly correlated with a competing class (the class other than the intended target). Table-\ref{tab:score_table} highlights the high correlation scores between the universal triggers and their associated dominant (or majority) class. When these triggers are appended to SNLI examples from the targeted class, they exploit the model's reliance on spurious correlations, leading it to favor the competing class over the intended target. For example, the universal trigger \textit{nobody} is closely associated with the contradiction class. When prepended to examples from the entailment class, it causes the model to misclassify them as contradictions (1st row in Table-\ref{tab:universal_trigger_effect}).

\subsubsection{De-biasing the Model}
The uniform distribution of universal triggers in the Trigger-Augmented dataset helps the model unlearn spurious correlations present in the original SNLI dataset. The two-stage training process balances foundational learning from the original data while mitigating biases using the Trigger-Augmented dataset. As shown in Table-\ref{tab:summary} (1st, 2nd and, 5th row), this approach significantly enhances the model’s overall performance and resilience (as highlighted in Section-\ref{ssec:Inoculation}).

\subsubsection{Decoding Attacks on the Contradiction Class} 
The contradiction class contains more correlated words in comparison to the entailment and neutral classes. The cumulative frequency of the top five correlated words is 312 for contradictions, 128 for neutral, and 57 for entailment. This abundance of correlated words makes contradictions particularly vulnerable. However, flipping predictions for contradiction-class examples to entailment or neutral by simply prepending tokens is feasible only if the example lacks these giveaway words. This is the reason why ELECTRA model's ability to correctly predict contradiction examples reduces by only 7.43\% with the introduction of universal triggers (refer to the results in Table-\ref{tab:summary}).

\section{Conclusion and Future work}
\label{sec:Conclusion}
In this study, we systematically investigated the vulnerabilities and biases in NLI models, proposing methods to enhance their robustness. Our findings demonstrated the effectiveness of universal triggers in exploiting spurious correlations to manipulate NLI model predictions, significantly outperforming random triggers. Moreover, Trigger-Augmented training proved successful in mitigating biases, thereby improving model resilience. This approach also underscored the nuanced challenges associated with attacking the contradiction class, shedding light on areas requiring further exploration.

For future work, we aim to explore diverse attack strategies \cite{song2021universaladversarialattacksnatural} beyond merely prepending triggers to hypotheses. Such strategies will help uncover additional weaknesses in NLI datasets, providing deeper insights into designing more robust datasets and improving the training processes for NLI models.

\bibliography{main}
\bibliographystyle{acl_natbib}

\end{document}